\documentclass[sigconf]{acmart}
\AtBeginDocument{%
  \providecommand\BibTeX{{%
    \normalfont B\kern-0.5em{\scshape i\kern-0.25em b}\kern-0.8em\TeX}}}

\usepackage{multirow} 
\usepackage{graphicx}
\usepackage{url}
\usepackage{float}
\usepackage{subfig}
\usepackage{overpic}  
\usepackage{bm}
\usepackage{amsmath}
\usepackage{amsfonts}
\usepackage{nicefrac}     
\usepackage{microtype}      
\usepackage{lipsum}
\usepackage{color}
\usepackage{booktabs}
\usepackage{epsfig}
\usepackage{epstopdf}
\usepackage{pifont}
\usepackage{graphicx}
\usepackage{array}
\usepackage[linesnumbered,ruled,vlined]{algorithm2e}

\newcommand{\name}{MUSE}

\setcopyright{acmcopyright}
\copyrightyear{2018}
\acmYear{2018}
\acmDOI{XXXXXXX.XXXXXXX}

\acmConference[Conference acronym 'XX]{Make sure to enter the correct
  conference title from your rights confirmation emai}{June 03--05,
  2018}{Woodstock, NY}
\acmPrice{15.00}
\acmISBN{978-1-4503-XXXX-X/18/06}

\begin{document}

\title{MUSE: Multi-View Contrastive Learning for Heterophilic Graphs}

\author{Mengyi Yuan}
\affiliation{%
  \department{School of Data Science and Engineering}
  \institution{East China Normal University}
  \city{Shanghai}
  \country{China}}
\email{mengyiyuan@stu.ecnu.edu.cn}

\author{Minjie Chen}
\affiliation{%
  \department{School of Data Science and Engineering}
  \institution{East China Normal University}
  \city{Shanghai}
  \country{China}}
\email{minjiechen@stu.ecnu.edu.cn}

\author{Xiang Li}
\authornote{Corresponding author.}
\affiliation{%
  \department{School of Data Science and Engineering}
  \institution{East China Normal University}
  \city{Shanghai}
  \country{China}}
\email{xiangli@dase.ecnu.edu.cn}

\renewcommand{\shortauthors}{Trovato and Tobin, et al.}

\begin{abstract}
In recent years, self-supervised learning has emerged as a promising approach in addressing the issues of label dependency and poor generalization performance in traditional GNNs. 
However, existing self-supervised methods have limited effectiveness on heterophilic graphs, due to the homophily assumption that results in similar node representations for connected nodes. 
In this work, we propose a multi-view contrastive learning model for heterophilic graphs, namely, \name.
Specifically,
we construct two views to capture the information of the ego node and its neighborhood by GNNs enhanced with contrastive learning, respectively. 
Then we integrate the information from these two views to fuse the node representations.
Fusion contrast is utilized to enhance the effectiveness of fused node representations.
Further, considering that the influence of neighboring contextual information on information fusion may vary across different ego nodes, we employ an information fusion controller to model the diversity of node-neighborhood similarity at both the local and global levels.
Finally, an alternating training scheme is adopted to ensure that unsupervised node representation learning and information fusion controller can mutually reinforce each other.
We conduct extensive experiments to
evaluate the performance of \name\ on 9 benchmark datasets. Our results show the effectiveness of \name\ on both node classification and clustering tasks.
We provide our data and codes at \url{https://anonymous.4open.science/r/MUSE-BD4B}.

\end{abstract}





\maketitle

\section{Introduction}

Graph neural networks (GNNs) \cite{kipf2016semi, velivckovic2017graph, gasteiger2018predict, wu2019simplifying} have shown remarkable success in learning representations of graph-structured data. 
By iteratively aggregating information from a node’s neighbors, 
GNNs map high-dimensional node representations into low-dimensional ones while preserving the useful information in the graph,
which can be further applied to a variety of downstream tasks, such as node classification \cite{kipf2016semi,gasteiger2018predict,xu2018representation} and node clustering \cite{wang2019attributed,zhang2019attributed}. 

\begin{figure}[t]
  \centering
  \includegraphics[width=\linewidth]{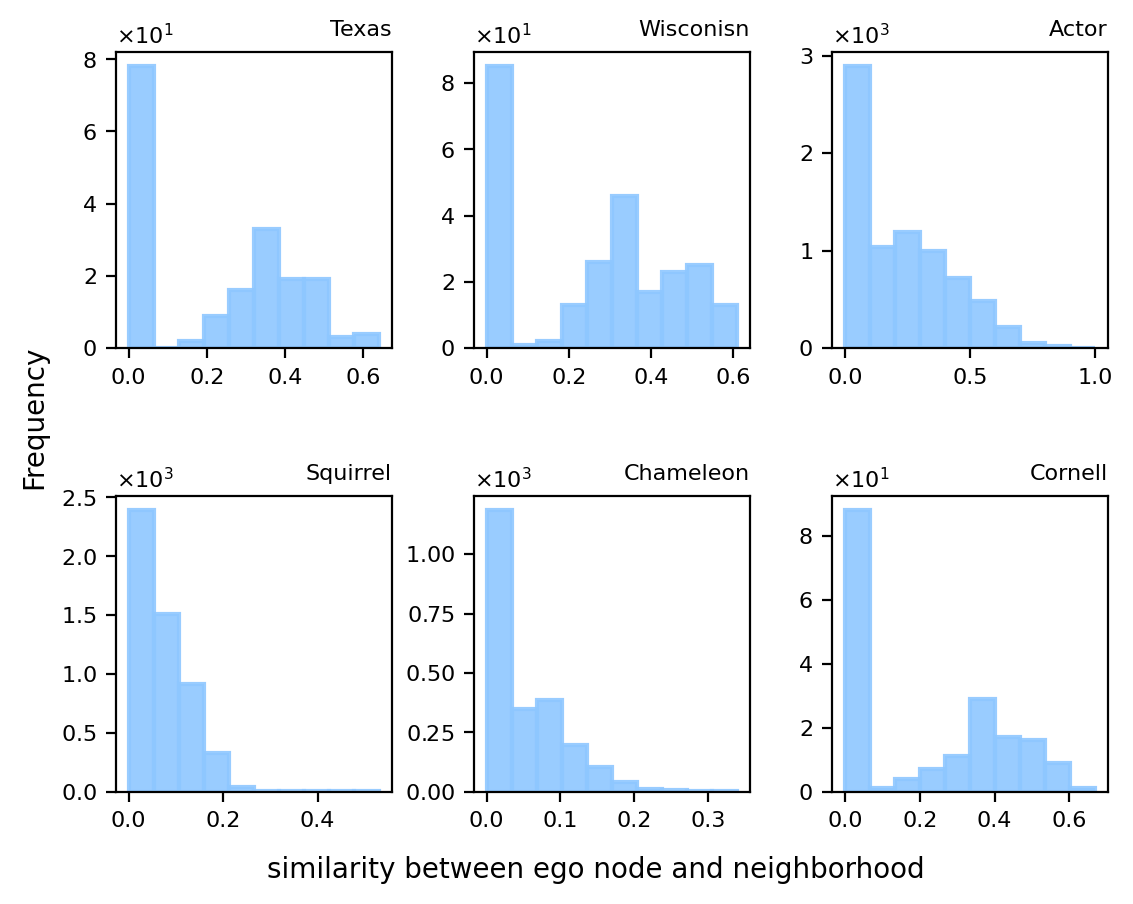}
  \caption{The distribution of similarity between ego node and its neighborhood in real-world graphs.}
  \Description{similarity distribution.}
  \label{img1}
\end{figure}

Traditional GNNs mainly adopt a semi-supervised approach, which still depends on labeled data and poses several problems, like label sparsity and poor transferring capacity \cite{bengio2012deep,sun2019infograph}. 
Recent works have shown that combining self-supervised learning with GNNs can improve the robustness and generalization ability of the model \cite{velickovic2019deep, peng2020graph, hassani2020contrastive, qiu2020gcc}, 
since graph self-supervised learning can
capture the underlying semantic and structure of the graph in an unsupervised manner.
However, existing graph self-supervised methods are mainly based on the homophily assumption, 
leading to similar node representations between ego node and its neighboring nodes. 
In real-life scenarios, the assumption of homophily often does not hold, and the performance of these methods will significantly deteriorate. We conduct similarity analysis between the ego nodes and their neighborhoods on six heterophilic graphs, as shown in Figure \ref{img1}. 
Evidently, there are a large number of extreme values at a low similarity level in each graph. This suggests that numerous nodes are dissimilar to their neighborhoods in real-world graphs.

Recently, most existing unsupervised learning methods \cite{tang2022graph, xiao2022decoupled, chen2022towards} on heterophilic graphs distinguish the information of the ego node and its neighborhood. 
However, some of them emphasize the information from neighborhood at the expense of retaining ego node information, resulting in even worse performance on certain datasets than MLP \cite{tang2022graph, xiao2022decoupled}. 
Some methods heavily rely on features of ego nodes to ensure their expressive power after multi-layer propagation in GNNs and fail to utilize the contextual information of neighboring nodes \cite{chen2022towards}.
These methods also perform poorly on certain datasets.
We attribute the poor generalization of these methods to the fact that they overlook the difference and diversity among nodes in a graph, 
which can also be observed from Figure \ref{img1}. 
From the perspective of an individual graph, 
there exists a large discrepancy on the similarity between different ego nodes and their neighborhoods, which reflects the local 
diversity among nodes in the graph.
We also observe that 
the global distribution of similarity among different graphs varies greatly.
To cope with these two types of diversities, 
we first divide the sources of information for nodes in a graph into two types: (1) features of ego node; 
(2) interaction between ego node and its neighborhood.
Then a question naturally arises: \emph{How to effectively fuse ego node information with neighboring contextual information to generate node representations?} 
On the one hand, 
information fusion needs to take into account the local difference of similarities between different ego nodes and their neighborhoods. 
On the other hand, effective representation fusion should reflect the global diversity across different graphs.

To address the aforementioned challenges, we propose a \textbf{MU}lti-view contra\textbf{S}tive learning method for h\textbf{E}terophilic graphs, namely, \textbf{\name}.
Our model consists of two major components: \emph{unsupervised node representation learning} and \emph{information fusion controller}. 
For the former, 
we construct two views, namely, 
\emph{semantic view} and \emph{contextual view}, 
based on which we 
capture relevant information with GNNs and employ contrastive learning to learn node representations in each view that are invariant to perturbations, respectively. 
After that, we fuse the embeddings learned from the two views to generate cross-view
node representations
with 
an information fusion controller.
This 
enables each node to adaptively integrate useful information from its neighborhood. 
To enhance the effectiveness of 
fused node representations,
we further introduce contrasive learning to drive them to be perturbation invariant.
For the information fusion controller,
it determines the fusion weight by
taking into account the diversity of node-neighborhood similarity at both the local and global levels,
thereby leading to personalized information fusion.
On the one hand,
we consider the local diversity by measuring the similarity between embeddings generated from both semantic and contextual views.
On the other hand, 
we handle the global diversity by constraining the distribution of similarity across the entire graph.
Considering that the two components mutually reinforce each other, we adopt an alternating training scheme to optimize them simultaneously.
Finally,
our main contributions can be summarized as follows:
\begin{itemize}
\item 
We propose \name, 
a novel contrastive learning model for heterophilic graphs.
To our best knowledge,
\name\ is the first heterophilous graph contrastive learning method that 
considers both local and global node similarity diversity.
\item We design an effective information fusion controller to model
the diversity of node-neighborhood similarity,
leading to personalized node representation fusion.

\item 
{We conduct extensive experiments on 9 benchmark datasets to evaluate the performance of \name.
Our results show its superiority over other state-of-the-art competitors.
}
\end{itemize}

\section{RELATED WORK}
\subsection{GNNs with heterophily}
Existing graph neural network methods on heterophilic graphs can mainly be divided into two categories. One is to capture information from distant nodes \cite{li2022finding, liu2021non, abu2019mixhop, pei2020geom, suresh2021breaking}.
For example,
MixHop \cite{abu2019mixhop} concats information from multi-hop neighbours at each GNN layer. Geom-GCN \cite{pei2020geom} discovers potential neighbors in a continuous latent space. Both the neighbors in the original graph latent space are aggregated during the message passing. WRGAT \cite{suresh2021breaking} captures the information from distant nodes by defining the type and weight of edges in the entire graph to reconstruct a computation graph.

The other is to adaptively aggregate useful information from the neighborhood by refining the GNN architecture \cite{chen2020simple, luan2021heterophily, yan2021two, zhu2020beyond, bo2021beyond}. 
For example,
H2GCN \cite{zhu2020beyond} excludes the self-loop connection and adopts an non-mixing operation in the GNN layer to emphasize the features of ego node. GGCN \cite{yan2021two} learns a weighted combination of the prior layer node representations, with signed weight denoting different neighbors at every layer of GNN. ACM \cite{luan2021heterophily} uses high-pass, low-pass and identity filter to aggregate neighbor information by different channels.

\subsection{Unsupervised graph representation learning }
Unsupervised graph representation learning methods can be divided into two types: generation-based methods and contrast-based methods.

Generation-based methods reconstruct the graph data from the perspectives of feature and structure of the graph, and use the input data as the supervision signal. Classic generation-based methods on graph include GAE \cite{kipf2016variational}, VGAE \cite{kipf2016variational}, SIG-VAE \cite{hasanzadeh2019semi} which focus on reconstructing the structure information of the graph and MGAE \cite{wang2017mgae}, GALA \cite{park2019symmetric} which put emphasis on the recontruction of the feature information of the graph.

Contrast-based methods construct representations under different views and maximize their agreement. According to the scale of contrast, existing methods come into three sub-categories: Node-to-node contrast, graph-to-graph contrast, and node-to-graph contrast. For example, GRACE \cite{zhu2020deep} pulls the representations of the same node closer under different augmentations and pushes away the representations of other nodes. GraphCL \cite{you2020graph} brings the graph-level representations closer under different views to ensure perturbation invariance. DGI \cite{velickovic2019deep} and MVGRL \cite{hassani2020contrastive} maximize the mutual information between node-level representations and graph-level representations, aiming to capture both local and global information.


However, most of these methods are based on the homophily assumption. Recent works have moved the emphasis to unsupervised scenarios on heterophilic graphs. Based on graph contrastive learning, HGRL \cite{chen2022towards} improves the node representations on heterophilic graphs by preserving the node original features and capturing the non-local neighbors. GREET \cite{liu2022beyond} discriminates homophilic edges from heterophilic edges and uses low-pass and high-pass filters to capture the corresponding information. Based on generation methods, DSSL \cite{xiao2022decoupled} decouples the diverse patterns in local neighborhood distribution to capture both homophilic and heterophilic information. NWR-GAE \cite{tang2022graph} emphasizes the role of the topological structure in the graphs and reconstructs the neighborhoods based on the local structure and features.

\section{PRELIMINARIES}
In this section, we introduce the notations used in this paper and  a brief background of GNNs.

\textbf{Notations}
Let $\mathcal{G}=(\mathcal{V}, \mathcal{E})$ be an undirected, unweighted graph, where $\mathcal{V}=\left\{v_1, \cdots, v_N\right\}$ is the set of nodes and $\mathcal{E} \subseteq \mathcal{V} \times \mathcal{V}$ is the set of edges. $\mathbf{X} \in \mathbb{R}^{N \times F}$ is the feature matrix where the $i$-th row $x_i$ is the $F$-dimensional feature vector of node $v_i$. $\mathbf{A} \in \mathbb{R}^{N \times N}$ denotes the binary adjacent matrix with $\mathbf{A_{i, j}}=1$ if $e_{i, j} \in \mathcal{E}$ and $\mathbf{A_{i, j}}=0$ otherwise. The neighboring set of node $v$ is denoted as $\mathcal{N}(v)$.

\textbf{Graph Neural Networks}
GNNs adopt a message passing mechanism, where the representation of each node $v\in\mathcal{V}$ is updated by aggregating messages from its local neighboring nodes, and then combining the aggregated messages with the node's own representation. Generally, given a GNN model $f(\cdot)$, message passing in the $l$-th layer can be divided into two operations: one is to aggregate information from a node's neighbors while the other is to update a node's representation. 
Given a node $v_i$, these two operations are formulated as:
\begin{equation}
  m_{{i}}^{(l)} = \texttt{AGGREGATE}^{(l)}\{{h_{{j}}^{(l-1)}, \forall v_{j} \in \mathcal{N}(v_{i})}\},
\end{equation}
\begin{equation}
  h_{{i}}^{(l)} = \texttt{COMBINE}^{(l)} \{ h_{{i}}^{(l-1)}, m_{{i}}^{(l)} \},
\end{equation}
where $m_{{i}}^{(l)}$ and $h_{{i}}^{(l)}$ denote the message vector and representation of node $v_{i}$ in the $l$-th layer, respectively. 
\texttt{AGGREGATE}$^{(l)}(\cdot)$ and \texttt{COMBINE}$^{(l)}(\cdot)$ are two functions in each GNN layer. 
\begin{figure*}[t]
  \centering
  \includegraphics[width=\textwidth]{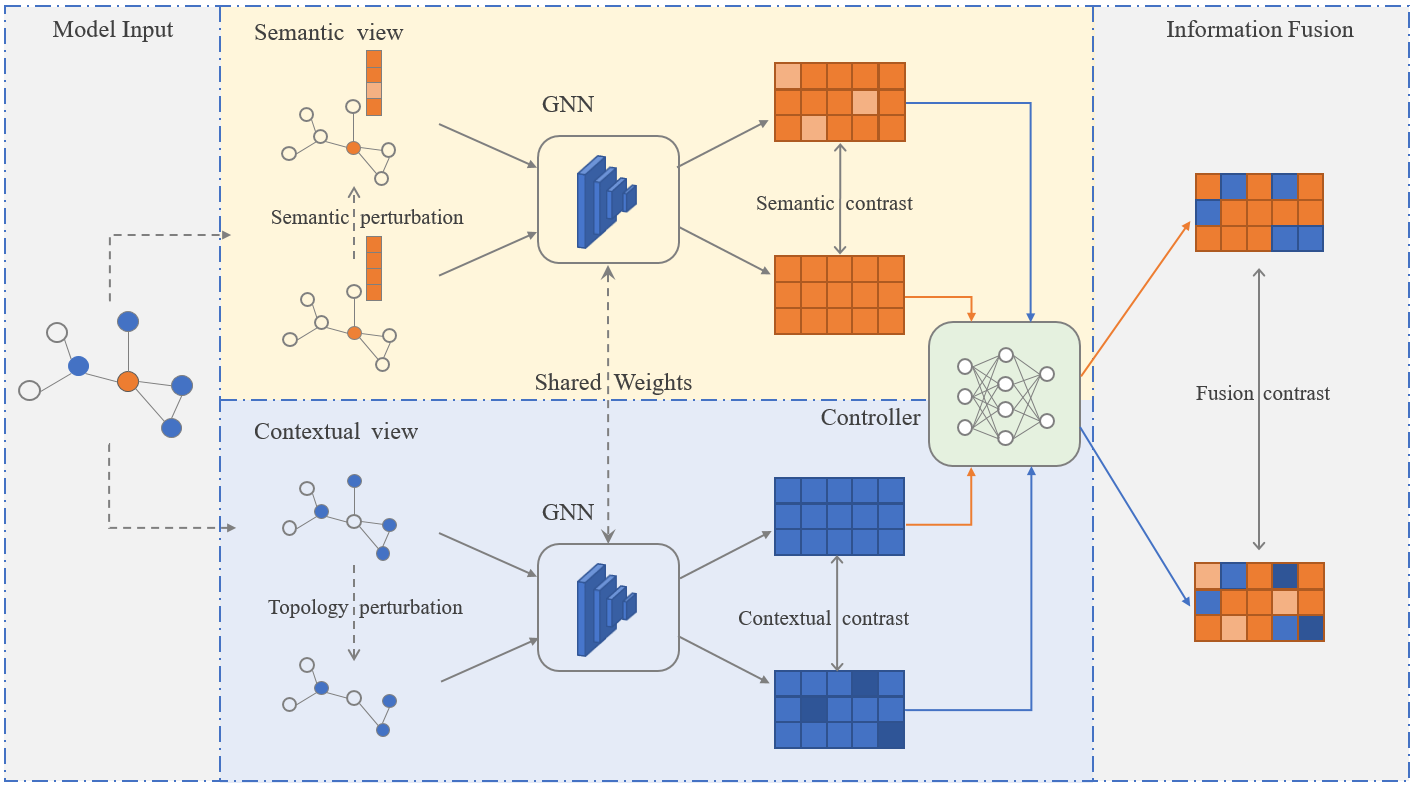}
  \caption{The overall framework of the proposed \name~ method.}
  \Description{framework}
  \label{framework}
\end{figure*}

\section{METHODOLOGIES}
In this section, we introduce the model design of our proposed \name~ method which is illustrated in Figure \ref{framework}. 
Our approach consists of two components: unsupervised node representation learning and information fusion controller.
\name~ first constructs two views to capture the semantic and contextual information of nodes in a graph with GNN. 
Semantic and contextual contrast are served as supervision signals to learn relevant node representations invariant to perturbations. 
\name~ then sends the information captured under two views into the information fusion controller to model the diversity of similarity between two kinds of information and generate the fused node representations in a node-specific way. 
Fusion contrast is employed to enhance the effectiveness of the fused node representations.
Since the two components in \name~ have a mutually reinforcing effect, an alternating training scheme is adopted to optimize these two components simultaneously.

\subsection{View Construction}
In heterophilic graphs, ego node and its neighborhood often exhibit high degree of heterophily due to differences in features and structure. To address this issue, we adopt a node-specific approach and construct two views: the semantic and the contextual view, for ego node and its neighborhood, respectively.
The semantic view describes the nodes with their intrinsic properties, 
while the contextual view characterizes nodes based on their local neighborhoods.

\subsubsection{Semantic View} Under semantic view, nodes that represent similar features in the graph are considered similar. Semantic-level contrast is employed as a form of supervision signal, aiming to encourage the learned representations of nodes with similar features to be consistent. For the set of initial features of nodes in the graph, we employ a perturbation $\mathbf{\tau_\alpha}$ to generate a new set of features as positive samples:
\begin{equation}
\left\{\widetilde{\mathbf{x}_1}, \widetilde{\mathbf{x}_2} \ldots, \widetilde{\mathbf{x}_N}\right\} \sim \mathbf{\tau_\alpha}\left(\left\{\mathbf{x}_1, \mathbf{x}_2, \ldots, \mathbf{x}_N\right\}\right),
\end{equation}
where $\widetilde{\mathbf{x}_i}$ is the augmented sample of $\mathbf{x}_i$.

We apply perturbations by altering only the features of the ego nodes while keeping the structure of graph unchanged.
Specifically, we randomly mask the initial node features in different dimensionality with a probability of $p_s$. We sample a binary vector $\mathbf{m}\in \mathbb{R}^{1 \times F}$ from the Bernoulli distribution with a probability of $(1-p_s)$, i.e., $m_i\sim Bernoulli\left(1-p_s\right)$, $i\in\left\{1, \cdots, F\right\}$, and perform element-wise multiplication with the features of each node:
 \begin{equation}
 \widetilde{\mathbf{x}_i}=\mathbf{x}_i\circ\mathbf{m},
\end{equation}

The initial features are fed into the GNN encoder $f(\cdot): \mathbb{R}^{N \times F} \rightarrow \mathbb{R}^{N \times F^{\prime}}$ to capture the semantic information of each node,
where $F^{\prime}$ is the embedding dimensionality. To independently encode each node without taking into account the information from neighboring nodes, we construct a unit matrix $\mathbf{I}\in \mathbb{R}^{N \times N}$ as the adjacency matrix and feed it into $f(\cdot)$ along with initial features. The semantic representations obtained are denoted as $\mathbf{H}^{s}\in \mathbb{R}^{N \times F^{\prime}}$:
\begin{equation}
\mathbf{H}^{s}=f_{\omega}\left(\mathbf{X}, \mathbf{I}\right),
\end{equation}
where $\omega$ denotes the parameters of GNN encoder.

 Given a node $v_i$, the semantic representation of $x_i$, denoted by $\mathbf{h}_i^{s}$, is the anchor sample. The semantic representation of the corresponding augmentation sample $\widetilde{\mathbf{x}_i}$ , denoted by $\widetilde{\mathbf{h}_i^{s}}$,  is treated as a positive sample of $\mathbf{h}_i^{s}$. The node representations of other nodes in the graph and their corresponding augmentations are considered as the negative samples of $\mathbf{h}_i^{s}$.

 We construct the semantic-level contrastive loss based on the normalized
temperature-scaled cross entropy loss (NT-Xent) \cite{chen2020simpleframe}. 
Firstly, a non-linear projection $g(\cdot): \mathbb{R}^{N \times F^{\prime}} \rightarrow \mathbb{R}^{N \times F_p}$ is used to map the node representations to a latent space where contrastive loss is applied. $F_p$ is the dimensionality of projected representations. The non-linear projection head $g$ is verified to be able to remove the information related to the downstream tasks from the node representations \cite{chen2020simpleframe}. Then we measure the similarity between pairs of samples in this latent space. Measuring metric between sample pair $\left(\mathbf{h}_i, \mathbf{h}_j\right)$ is defined as $\theta\left(\mathbf{h}_i, \mathbf{h}_j\right)=s\left(g\left(\mathbf{h}_i, \mathbf{h}_j\right)\right)$, where $s$ denotes cosine similarity and $g$ is implemented with a two-layer MLP. Given a positive node pair $\left(\mathbf{h}_i^{s}, \widetilde{\mathbf{h}_i^{s}}\right)$, the pairwise semantic contrastive loss is formulated as:
\begin{equation}
\ell_s\left(\mathbf{h}_i^{s}, \widetilde{\mathbf{h}_i^{s}}\right)=-\log \frac{e^{\theta\left(\mathbf{h}_i^{s}, \widetilde{\mathbf{h}_i^{s}}\right) / \tau}}{\sum_{v_j \in \mathcal{V}\backslash v_i}e^{\theta\left(\mathbf{h}_i^{s}, \mathbf{h}_j^{s}\right) / \tau}+\sum_{v_j \in \mathcal{V}} e^{\theta\left(\mathbf{h}_i^{s}, \widetilde{\mathbf{h}_j^{s}}\right) / \tau}},
\end{equation}

Similarly, take $\widetilde{\mathbf{h}_i^{s}}$ as the anchor sample, $\mathbf{h}_i^{s}$ is the positive sample of $\widetilde{\mathbf{h}_i^{s}}$, and the negative samples remain unchanged. Correspondingly, pairwise contrastive loss between $\left(\widetilde{\mathbf{h}_i^{s}}, \mathbf{h}_i^{s}\right)$ is $\ell\left(\widetilde{\mathbf{h}_i^{s}}, \mathbf{h}_i^{s}\right)$, and we can obtain the semantic contrastive loss $\mathcal{L}_s$ as:
\begin{equation}
\mathcal{L}_{s}=\frac{1}{2 N} \sum_{i=1}^N\left[\ell\left(\mathbf{h}_i^{s}, \widetilde{\mathbf{h}_i^{s}}\right)+\ell\left(\widetilde{\mathbf{h}_i^{s}}, \mathbf{h}_i^{s}\right)\right],
\end{equation}

\subsubsection{Contextual View}  Contextual view is constructed to capture the information from the interaction between ego node and its neighborhood. Different from the semantic view, here we focus on the contextual information provided by neighboring nodes rather than the features of the ego node.

Under the contextual view,
nodes with similar local neighborhoods are considered to be similar.
Context-level contrast is further employed to keep the representations of nodes with similar contexts to be consistent. 
When constructing positive samples under the contextual view, 
it should be ensured that the semantic information of nodes remains unchanged. 
Therefore, we introduce perturbations $\mathbf{\tau_\beta}$ to the topology of the neighboring nodes while preserving their semantic information:
\begin{equation}
\widetilde{\mathbf{A}} \sim \mathbf{\tau_\beta}\left(\mathbf{A}\right),
\end{equation}
In practice, we randomly drop some edges in the graph with a probability of $p_c$. This is equivalent to removing a small portion of nodes from the neighborhood of each ego node to alter the contextual information.
Specifically,
we sample a binary masking matrix $\mathbf{E}\in \left\{0,1\right\}^{N \times N}$ from the Bernoulli distribution with a probability of $(1-p_c)$, i.e., $e_{ij}\sim Bernoulli\left(1-p_c\right)$, $i,j\in\left\{1, \cdots, N\right\}$, and perform element-wise multiplication with the adjacent matrix:
\begin{equation}
\widetilde{\mathbf{A}} = \mathbf{A}\circ\mathbf{E},
\end{equation}
To capture the contextual information of each node, 
we use {feature matrix and adjacency matrix} as the input of GNN encoder $f(\cdot)$ and obtain the contextual representations denoted as $\mathbf{H}^{c}\in \mathbb{R}^{N \times F^{\prime}}$ by:
$$
\mathbf{H}^{c}=f_{\omega}\left(\mathbf{X}, \mathbf{A}\right).
$$

Similar as the contrastive loss under the semantic view, we construct the positive sample pair $\left(\mathbf{h}_i^{c}, \widetilde{\mathbf{h}_i^{c}}\right)$ and the corresponding negative sample pairs. The contextual contrastive loss is defined as:
\begin{equation}
\mathcal{L}_{c}=\frac{1}{2 N} \sum_{i=1}^N\left[\ell\left(\mathbf{h}_i^{c}, \widetilde{\mathbf{h}_i^{c}}\right)+\ell\left(\widetilde{\mathbf{h}_i^{c}}, \mathbf{h}_i^{c}\right)\right].
\end{equation}

\subsection{Cross-view Node Representation Fusion}

As previously discussed, we divide the source of information for a node in the graph into two types: (1) the node's own features, (2) the information brought by the interaction between the ego node and its neighborhood.
To further enhance node representation learning in a graph, 
we fuse node embeddings learned from both the semantic view and the contextual view.
We consider that the contextual information plays a complementary role for the semantic information of the ego node.
This fusion process not only captures the intrinsic properties of an individual ego node, but also includes the features and structural information of its neighborhood, thereby providing a more holistic node representation.

However, in practice, the amount of contextual information that each ego node needs varies, so we need to perform node-specific cross-view embedding fusion. Specifically, given a node $v_i$, 
we represent its fused representation 
as follows :
\begin{equation}
h_i^f=h^s_i+\lambda_i h^c_i,
\end{equation}
where $\lambda_i$ is a personalized weight which we will introduce in the next section.

Considering that the fused node representations should still remain invariant to perturbations, we introduce the fusion contrast to the unsupervised node representation learning.
Given $\mathbf{h}_i^{f} = \mathbf{h}_i^{s}+\mathbf{\lambda}_i\mathbf{h}_i^{c}$ as an anchor sample, we consider $\widetilde{\mathbf{h}_i^{f}}=\widetilde{\mathbf{h}_i^{s}}+\mathbf{\lambda}_i\widetilde{\mathbf{h}_i^{c}}$ as the positive sample. We  minimize the distance between the fused node representations before and after perturbations to ensure the invariance to perturbations. The cross-view fusion contrastive loss is defined as:
\begin{equation}
\mathcal{L}_{f}=\frac{1}{2 N} \sum_{i=1}^N\left[\ell\left(\mathbf{h}_i^{f}, \widetilde{\mathbf{h}_i^{f}}\right)+\ell\left(\widetilde{\mathbf{h}_i^{f}}. \mathbf{h}_i^{f}\right)\right],
\end{equation}
To sum up, the unsupervised node representation loss can be obtained as:
\begin{equation}
\mathcal{L}_{contrast} =\mathcal{L}_{s}+\beta_1\mathcal{L}_{c}+\beta_2 \mathcal{L}_{f}\label{contrast},
\end{equation}
where $\beta_1$ and $\beta_2$ are weight factors for adjusting the importance of different components.

\subsection{Information Fusion Controller}

Considering the diversity across different nodes, we propose an information fusion controller $\psi$ to model the diversity of similarity between the information deprived from two views and allow each node to adaptively integrate two kinds of information.
Since node embeddings learned from the two views have a direct impact on the information fusion, we first filter the noisy features in $h^s$ and $h^c$.
Specifically,
for node $v_i$,
we have:
\begin{equation}
w_i^s = \varphi(h_i^s; \phi_1), w_i^c = \varphi(h_i^c; \phi_2),
\end{equation} 
where we implement the filter $\varphi$ with a one-layer MLP and the output of filter is $w^s\in{R}^{N \times F_g}$, $w^c\in{R}^{N \times F_g}$.

Given a node $v_i$, except the node's semantic properties and the contextual characteristics of its neighbors, the information fusion controller further takes into account the degree centrality. This measures the position of a node within the structure of the graph, which has been evidenced to benefit the performance of GNN on heterophilic graphs \cite{ma2021homophily}. 
The degree centrality is formulated as:
\begin{equation}
d_i=\sum_{j=1}^{N}\mathbf{A}_{ij},
\end{equation}
We calculate the personalized weight factor $\lambda_i$ for each node $v_i$ {with a two-layer MLP} by:
\begin{equation}
\lambda_{i} = \psi\left(w_{i}^{s}, w_{i}^{c}, \mathbf d_i; \phi_3\right),\label{lam}
\end{equation}

As shown in Figure \ref{img1}, from the node level, 
there typically exists difference in the similarity between different ego nodes and their neighborhoods. When a node's semantic representation and contextual representation share high similarity, we consider that the contextual information provided by its neighboring nodes heavily overlaps with the node's own information. 
Therefore, this node needs less contextual information from neighborhood.
From the graph level, the distribution of similarity between ego nodes and their neighborhoods in the whole graph varies across different graphs. 
Therefore, 
we impose some constraints on the overall distribution of $\lambda$ with two regularization terms. Specifically, we constrain
the average $\lambda$ value 
for each individual graph to be a pre-set hyper-parameter. Additionally, considering that the overall distribution of all nodes in each individual graph may vary significantly, we use $L_2$ norm to {restrict} the magnitude of this diversity. 
Therefore, we have the following objective function:
\begin{equation}
\mathcal{L}_\phi = \sum_{i=1}^N\lambda_{i} s\left({h}_{i}^{s}, {h}_{i}^{c}\right) + \alpha_1 ||\lambda||_2\label{phi} +\alpha_2 \left|\frac{1}{N}\sum_{i=1}^{N}\lambda_i-\epsilon\right|,
\end{equation} 
where $s(\cdot)$ denotes cosine similarity, coefficients $\alpha_1$ and $\alpha_2$ represent the weight factors in the objective, {$\epsilon$ is a hyper-parameter that controls the average $\lambda$ in the graph,} $\phi$ denotes all the parameters in information fusion controller.

\subsection{Overall Framework}
\subsubsection{Model Training} 
From the discussion above, we can obtain the overall optimization objective of \name~ as:
\begin{equation}
\mathcal{L}_{total} = \mathcal{L}_{contrast}+\mathcal{L}_\phi.
\end{equation}
\name~ can be divided into two parts: unsupervised node representation learning and information fusion controller. 

In the first part, we apply attribute-level perturbation to the initial input under the semantic view, and feed it into GNN to obtain corresponding node representations $\mathbf{H}^s$ and $\mathbf{\widetilde{H^s}}$.
Similarly, topology-level perturbation is applied to the structure of the graph and we can obtain the contextual node representations, denoted as $\mathbf{H}^c$ and $\mathbf{\widetilde{H^c}}$. 
Furthermore, we fuse the node representations as $\mathbf{H}^{f} = \mathbf{H}^{s}+\mathbf{\lambda}\mathbf{H}^{c}$ to integrate information from two views.

In the second part, the semantic node representations $\mathbf{H}^s$ and contextual node representations $\mathbf{H}^c$ obtained in the first part is sent into an information fusion controller $\psi$ to model the diversity of two kinds of information. The personalized weight factor $\mathbf{\lambda}$ learned by the information fusion controller fuses the node representations in a node-specific way.

\subsubsection{Optimization Strategy} The two components in our model are coupled with each other. On the one hand, unsupervised node representation learning relies on the information fusion controller to control the integration of information from semantic and contextual views, and further generate node representations. 
On the other hand, the information fusion controller {models the diversity of node-neighborhood similarity} with node representations derived from different views. 
Therefore, we adopt an alternating training strategy to ensure the effectiveness of both components simultaneously.
Specifically,
during training, we first fix the information fusion controller $\psi_{\phi}\left(\cdot\right)$ and train the encoder $f_\omega\left(\cdot\right)$ and projector $g_{\mu}\left(\cdot\right)$ by back-propagating $\mathcal{L}_{contrast}$ to learn node representations. 
Then, encoder $f_\omega\left(\cdot\right)$ and projector $g_{\mu}\left(\cdot\right)$ are fixed. 
Semantic and contextual node representations generated by the encoder $f_\omega\left(\cdot\right)$ are sent into the information fusion controller $\psi_{\phi}\left(\cdot\right)$. We compute the controller loss $\mathcal{L}_\phi$ to optimize the information fusion controller $\psi_{\phi}\left(\cdot\right)$. 
Here, $\omega$ represents the parameters of encoder $f\left(\cdot\right)$, $\mu$ represents the parameters of projector $g\left(\cdot\right)$, $\phi$ denotes the parameters of information fusion controller $\psi\left(\cdot\right)$.
It is worth noting that at the first epoch, $\mathbf{\lambda}$ used to update the GNN parameters via $\mathcal{L}_{contrast}$ is obtained from the information fusion controller that is randomly initialized.

\subsubsection{Time Complexity Analysis}
We next analyze the time complexity of the two main components in our model. 
In unsupervised node representation learning, the time complexities of GNN encoder and projection head are $\mathcal{O}(F|\mathcal{E}|L+NFF^{\prime}L)$ and $\mathcal{O}(NF^{\prime}F_{p}L_{p})$, where $|\mathcal{E}|$ is the number of edges, $L$ and $L_p$ are the layers of GNN and projection head, respectively, {$F$ is the dimensionality of initial features, $F^{\prime}$ is the dimensionality of final node representations, $F_{p}$ is the dimensionality of projected representations in projection head and $N$ is the number of nodes in a graph.} 
Further,
for the 
information fusion controller,
the time complexities of the noise filter
and computing $\lambda$
in Equation~\ref{lam}
are 
$\mathcal{O}(NF^{\prime}F_{g})$
and 
$\mathcal{O}(NF_{g})$, respectively, 
where $F_{g}$ is the output dimensionality of the filter and $F_{g}, F^{\prime} \ll N$.

\section{Experiments}
\subsection{Experimental Settings}
\subsubsection{Datasets} We evaluate our model on nine real-world datasets, including three homophilic datasets (Cora, CiteSeer, PubMed \cite{sen2008collective}) and six heterophilic datasets (Cornell, Texas, Wisconsin, Chameleon, Squirrel, Actor \cite{pei2020geom}). Among them, Cora, Citeseer, Pubmed are citation networks. Cornell, Texas, and Wisconsin are school department webpage networks. Chameleon and Squirrel are Wikipedia networks. Actor is an actor co-occurrence network in Wiki pages. Details of these datasets are summarized in the Table \ref{dataset}. 

\begin{table}[H]
  \caption{Datasets statistics.}
  \label{dataset}
  \begin{tabular}{ccccc}
    \toprule
    \toprule
    Datasets&Node&Edges&Features&Classes\\
    \midrule
    Cornell & 183 & 295 & 1,703 & 5\\
    Texas & 183 & 309 & 1,703 & 5\\
    Wisconsin & 251 & 499 & 1,703 & 5\\
    Chameleon & 2,277 & 36,051 & 2325 & 5\\
    Squirrel & 5,201 & 216,933 & 2089 & 5\\
    Actor & 7,600 & 29,926 & 932 & 5\\
    \midrule
    Cora & 2708 & 10,556 & 1,433 &7\\
    CiteSeer & 3,327 & 9,104 & 3,703 & 6\\
    PubMed & 19,717 & 88,648 &500 &3\\
  \bottomrule
  \bottomrule
\end{tabular}
\end{table}
\subsubsection{Baselines.}

(1) For node classification as the downstream task, we compare our model with four groups of baseline methods: supervised learning methods (i.e., GCN \cite{kipf2016semi}, GAT \cite{velivckovic2017graph}, and MLP), supervised methods specially designed for heterophilic graphs (i.e., WRGAT \cite{suresh2021breaking}, H2GCN \cite{zhu2020beyond}), contrast-based unsupervised learning methods designed for homophilic graphs (i.e., DGI \cite{velickovic2019deep}, GMI \cite{peng2020graph}, MVGRL \cite{hassani2020contrastive}, BGRL \cite{thakoor2021large}, GRACE \cite{zhu2020deep}) and unsupervised learning methods designed for heterophilic graphs (i.e., DSSL \cite{xiao2022decoupled}, NWR-GAE \cite{tang2022graph}, HGRL \cite{chen2022towards}, GREET \cite{liu2022beyond}).

(2) For node clustering as the downstream task, we compare our model with four groups of baseline methods: traditional unsupervised clustering methods (i.e., AE \cite{hinton2006reducing},  node2vec \cite{grover2016node2vec}, struc2vec \cite{ribeiro2017struc2vec}, LINE \cite{tang2015line}), attributed graph clustering methods (i.e., GAE(VGAE) \cite{kipf2016variational}, GraphSAGE \cite{hamilton2017inductive}, SDCN \cite{bo2020structural}), contrast-based unsupervised methods designed for homophilic graphs (i.e., MVGRL \cite{hassani2020contrastive},  GRACE \cite{zhu2020deep}, BGRL \cite{thakoor2021large}) and unsupervised methods designed for heterophilic graphs (i.e., DSSL \cite{xiao2022decoupled}, HGRL \cite{chen2022towards}).

\subsubsection{Implementation Details.} We implement our model by PyTorch and optimize the model by Adam optimizer \cite{kingma2014adam}.
We utilize a two-layer GCN \cite{kipf2016semi} as the GNN encoder, and conduct all the experiments following the standard linear evaluation scheme that is widely adopted \cite{velickovic2019deep, zhu2020deep, liu2022beyond}. In the training step, we train the model in an unsupervised learning manner to learn the node representations. 
In the evaluating step, the learned node representations are sent into the downstream tasks. 
We run the experiments with $10$ random splits, and report mean classification accuracy with standard deviation. 
We set $\alpha_2$ in Equation \ref{phi} to 1 and the embedding dimensionality of filter $F_g$ to 30. 
We fine-tune the following hyper-parameters:
$lr \in \{0.0001, 0.0005, 0.001, 0.005, 0.01\}$,
$lr_{controller} \in \{0.0001, 0.0005, 0.001, 0.005, 0.01\}$,
$\beta_1 \in \{0.001, 0.01, 0.1, 1, 10, 100\}$,
$\beta_2 \in \{0.001, 0.01, 0.1, 1, 10, 100\}$,
$p_s \in \{0.1, 0.2, 0.3, 0.4, 0.5\}$,
$p_c \in \{0.1, 0.2, 0.3, 0.4, 0.5\}$, 
$\epsilon \in \{0, 0.1, 0.2, \dots, 1\}$, 
$\alpha_1 \in \{10^2, 10^4, 10^6\}$,
$dropout \in \{0.1, 0.2, 0.3, 0.4\}$.

Two downstream tasks are employed to evaluate the effectiveness and  generalizability of the learned node representations. 
(1) Node classification: For homophilic graphs, we adopt the public splits with 20 nodes per class for training, 500 nodes for validation and 1,000 nodes for testing \cite{yang2016revisiting, kipf2016semi}. 
For heterophilic graphs, we adopt the commonly used training/validation/test split ratio of 48/32/20 as previous works \cite{pei2020geom, liu2022beyond}. 
A linear model is trained on top of the frozen node representations, and test accuracy is adopted as the evaluation metric. 
Results for GCN, GAT, MLP, DGI, GMI, MVGRL, BGRL, GRACE are reported from \cite{liu2022beyond}. For WRGAT, H2GCN, DSSL, NWR-GAE and GREET, results are derived from the original papers. 
For HGRL and results not reported in the original papers, we run their public code on standard splits while keeping other settings the same.
(2) Node clustering: Frozen node representation is fed into a K-means clustering model and the number of clusters is set as the number of classes. 
We adopt three evaluation metrics:  accuracy (ACC), normalized mutual information (NMI) and average rand index (ARI). 
We run the public code of DSSL and fine-tune its hyper-parameters to report the best results. Results for other baselines are directly derived from \cite{chen2022towards}.


\subsection{Experimental Results}\label{result}
\subsubsection{Node Classification Performance.}
Table \ref{classification} summarizes the performance results of node classification on three homophilic datasets and six heterophilic datasets. From the table, we make the following observations:

\begin{table*}[t]\small
  \caption{Results of node classification (in percent $\pm$ standard deviation). $\ast$ indicates the results are derived from the original papers. The best and runner-up results are highlighted with \textbf{bold} and \underline{underline}, respectively.}
  \label{classification}
  \setlength{\tabcolsep}{1mm}{
  \begin{tabular}{lcccccc|ccc|ccc}
    \toprule
    \toprule
    \multirow{2}{*}{Methods}&\multicolumn{6}{c|}{Heterophilic} & \multicolumn{3}{c|}{Homophilic} & \multicolumn{3}{c}{Average Performance}\\
    \cline{2-13}& Cornell & Texas & Wisconsin & Chameleon & Squirrel & Actor &Cora & CiteSeer & PubMed& All&Hete.&Homo.\\
    \midrule
    GCN & 57.03$\pm$3.30 & 60.00$\pm$4.80 & 56.47$\pm$6.55 & 59.63$\pm$2.32 & 36.28$\pm$1.52 & 30.83$\pm$0.77&81.5$^\ast$ & 70.3$^\ast$ & 79.0$^\ast$&59.00&50.04& 76.93 \\
    GAT & 59.46$\pm$3.63 & 61.62$\pm$3.78 & 54.71$\pm$6.87 & 56.38$\pm$2.19 & 32.09$\pm$3.27 & 28.06$\pm$1.48&83.0$^\ast$ & 72.5$^\ast$ & 79.0$^\ast$&58.54&48.72&78.17\\
    MLP & 81.08$\pm$7.93 & 81.62$\pm$5.51 & 84.31$\pm$3.40 & 46.91$\pm$2.15 & 29.28$\pm$1.33 & 35.66$\pm$0.94&56.11$\pm$0.34 & 56.91$\pm$0.42 & 71.35$\pm$0.05& 60.36& 59.81 &61.46 \\
    \midrule
    WRGAT$^\ast$ & 81.62$\pm$3.90 & 83.62$\pm$5.50 & \underline{86.98$\pm$3.78} & 65.24$\pm$0.87 &48.85$\pm$0.78 & 36.53$\pm$0.77&\textbf{88.20$\pm$2.26}&\underline{76.81$\pm$1.89}&\underline{88.52$\pm$0.92}&\underline{72.93}&\underline{67.14}&\underline{84.51}\\
    H2GCN$^\ast$ & 82.16$\pm$4.80 & 84.86$\pm$6.77 & 86.67$\pm$4.69 & 59.39$\pm$1.98 & 37.90$\pm$2.02 & 35.86$\pm$1.03&\underline{87.81$\pm$1.35}&\textbf{77.07$\pm$1.64}&\textbf{89.59$\pm$0.33}&71.26&64.47&\textbf{84.82}\\
    \midrule
    DGI & 63.35$\pm$4.61 & 60.59$\pm$7.56 & 55.41$\pm$5.96 & 39.95$\pm$1.75 & 31.80$\pm$0.77 & 29.82$\pm$0.69&82.29$\pm$0.56 & 71.49$\pm$0.14 & 77.43$\pm$0.84& 56.90&46.82 & 77.07\\
    GMI & 54.76$\pm$5.06 & 50.49$\pm$2.21 & 45.98$\pm$2.76 & 46.97$\pm$3.43 & 30.11$\pm$1.92 & 30.11$\pm$1.92 &82.51$\pm$1.47 & 71.56$\pm$0.56 & 79.83$\pm$0.90&54.45&42.69&77.97\\
    MVGRL & 64.30$\pm$5.43 & 62.38$\pm$5.61 & 62.37$\pm$4.32 & 51.07$\pm$2.68 & 35.47$\pm$1.29 & 30.02$\pm$0.70 &  83.03$\pm$0.27 & 72.75$\pm$0.46 & 79.63$\pm$0.38&60.11 &50.94&78.47\\
    BGRL & 57.30$\pm$5.51 & 59.19$\pm$5.85 & 52.35$\pm$4.12 & 47.46$\pm$2.74 & 32.64$\pm$0.78 & 29.86$\pm$0.75&  81.08$\pm$0.17 & 71.59$\pm$0.42 & 79.97$\pm$0.36&56.83 &46.47 &77.55\\
    GRACE & 54.86$\pm$6.95 & 57.57$\pm$5.68 & 50.00$\pm$5.83 & 48.05$\pm$1.81 & 31.33$\pm$1.22 & 29.01$\pm$0.78 &  80.08$\pm$0.53 & 71.41$\pm$0.38 & 80.15$\pm$0.34& 55.83&45.14&77.21\\
    \midrule
    DSSL & 53.15$\pm$1.28 & 62.11$\pm$1.53 & 56.29$\pm$4.42 &48.74$\pm$1.53 & 40.51$\pm$0.38 & 28.36$\pm$0.65&  83.06$\pm$0.53 & 73.51$\pm$0.64 & 82.98$\pm$0.49&58.75 &48.19&79.85\\
    NWR-GAE & 58.64$\pm$5.61 & 69.62$\pm$6.66 & 68.23$\pm$6.11 & \underline{72.04$\pm$2.59} & \textbf{64.81$\pm$1.83} &  30.17$\pm$0.17 &  83.62$\pm$1.61 & 71.45$\pm$2.41 & 83.44$\pm$0.92&66.89 &60.59&79.50\\
    HGRL & 79.46$\pm$4.45 & 82.16$\pm$6.00 & 86.28$\pm$3.58 & 48.29$\pm$1.64 & 35.79$\pm$0.89 & \underline{36.97$\pm$0.98} & 80.66$\pm$0.43 & 68.56$\pm$1.10 &  80.35$\pm$0.58&66.50 & 61.49 &76.52\\
    GREET & \textbf{85.14$\pm$4.87} & \underline{87.03$\pm$2.36} &84.90$\pm$4.48 &63.64$\pm$1.26 & 42.29$\pm$1.43 & 36.55$\pm$1.01&  83.81$\pm$0.87 & \underline{73.08$\pm$0.84} & 80.29$\pm$1.00&70.75 & 66.59 &79.06\\
    \midrule
    \name & \underline{82.16$\pm$3.42} & \textbf{89.73$\pm$2.79} & \textbf{88.24$\pm$3.20} & \textbf{72.37$\pm$2.21} & \underline{54.19$\pm$3.04} & \textbf{38.55$\pm$1.34} &82.24$\pm$0.41 & 71.14$\pm$0.40 & 82.90$\pm$0.59& \textbf{73.39}&\textbf{70.87}&78.76\\
    
  \bottomrule
  \bottomrule
\end{tabular}}
\end{table*}
\begin{table*}\footnotesize\centering
\caption{Results of node clustering (in percent $\pm$ standard deviation). The best and runner-up results are highlighted with \textbf{bold} and \underline{underline}, respectively.}
\label{cluster}
\setlength{\tabcolsep}{1mm}{
\begin{tabular}{lcccccccccccc}
\toprule
\toprule
\multirow{2}{*}{Methods}&\multicolumn{3}{c}{Texas} & \multicolumn{3}{c}{Actor} & \multicolumn{3}{c}{Cornell}& \multicolumn{3}{c}{CiteSeer}\\
\cline{2-13}
& ACC & NMI & ARI & ACC & NMI & ARI & ACC & NMI & ARI & ACC & NMI & ARI\\
\midrule
AE & 50.49$\pm$0.01& 16.63$\pm$0.01 &  14.60$\pm$0.01 &  24.19$\pm$0.11 &  0.97$\pm$0.03 & 0.50$\pm$0.04 & 52.19$\pm$0.01 & 17.08$\pm$0.01 & 17.41$\pm$0.01& 58.79$\pm$0.19 &30.91$\pm$0.21 & 30.29$\pm$0.23\\
node2vec& 48.80$\pm$1.93&2.58$\pm$0.70& -1.62$\pm$0.65 &25.02$\pm$0.04 &0.09$\pm$0.01& 0.06$\pm$0.02& 50.98$\pm$0.01 &5.84$\pm$0.01& 0.18$\pm$0.01 &20.76$\pm$0.27& 0.35$\pm$0.03& -0.01$\pm$0.04\\
struc2vec&49.73$\pm$0.01& 18.61$\pm$0.01& 20.97$\pm$0.01& 22.49$\pm$0.34& 0.04$\pm$0.01& -0.05$\pm$0.05 &32.68$\pm$0.01& 1.54$\pm$0.01& -2.20$\pm$0.01& 21.22$\pm$0.45& 1.18$\pm$0.08& 0.17$\pm$0.06\\
LINE &  49.40$\pm$2.08& 16.90$\pm$1.57& 18.08$\pm$1.06 &22.70$\pm$0.08 &0.09$\pm$0.01 &0.11$\pm$0.01 &34.10$\pm$0.77 &2.85$\pm$0.21 &-1.54$\pm$0.25& 28.42±0.88& 8.49$\pm$0.74 &3.54$\pm$0.56\\
\midrule
GAE & 42.02$\pm$1.22 &8.49$\pm$1.31& 10.83$\pm$1.92 &23.45$\pm$0.04 &0.18 $\pm$0.01 &-0.04$\pm$0.01& 43.72$\pm$1.25& 5.11$\pm$0.38& 6.51$\pm$1.74 &48.37$\pm$0.37& 24.59$\pm$0.22 &19.50$\pm$0.31\\
VGAE&  50.27$\pm$1.87& 11.73$\pm$0.95 &21.51$\pm$1.81 &23.30$\pm$0.22 &0.21$\pm$0.03 &0.34$\pm$0.05& 43.39$\pm$0.99& 5.46$\pm$0.46& 3.97$\pm$0.49 &55.67$\pm$0.13 &32.45$\pm$0.10 &28.34$\pm$0.13\\
GraphSAGE &  56.83$\pm$0.56 &16.97$\pm$1.93& 23.50$\pm$2.98& 23.08$\pm$0.29& 0.58$\pm$0.14& 0.22$\pm$0.07& 44.70$\pm$2.00 &4.33$\pm$0.93& 5.64$\pm$1.33& 49.28$\pm$1.18& 22.97$\pm$0.80& 19.21$\pm$1.33\\
SDCN & 44.04$\pm$0.56& 14.24$\pm$1.93& 10.65$\pm$2.98 &23.67$\pm$0.29 &0.08$\pm$0.14 &-0.01$\pm$0.07& 36.94$\pm$2.00& 6.6$\pm$0.93 &3.38$\pm$1.33& 59.86$\pm$1.18 &30.37$\pm$ 0.80& 29.70$\pm$1.33\\
\midrule
MVGRL &  \underline{62.79$\pm$2.33} &25.66$\pm$1.81& 33.54$\pm$4.59 &28.58$\pm$1.03 &2.42$\pm$0.51 &2.80$\pm$0.57 &43.77$\pm$3.03& 8.38$\pm$2.82 &7.09$\pm$2.95& 45.67$\pm$9.08 &23.41$\pm$7.73 &19.92$\pm$7.92\\
GRACE &  56.99$\pm$2.23 &20.65$\pm$1.02& 29.54$\pm$4.22 &25.87$\pm$0.45& 0.56$\pm$0.28& 0.93$\pm$0.38 &43.55$\pm$4.60& 8.23$\pm$1.16& 6.43$\pm$1.97 &54.66$\pm$5.41 &31.70$\pm$3.78 &27.40$\pm$5.60\\
BGRL & 58.68$\pm$1.80& 21.95$\pm$2.40& 23.74$\pm$2.29& 28.20$\pm$0.27& 1.84$\pm$0.19 &2.35$\pm$0.12& 55.08$\pm$1.68 &7.98$\pm$0.53 &3.92$\pm$1.03& \underline{64.27$\pm$1.68}& \underline{36.63$\pm$1.71}& \underline{36.71$\pm$1.85}\\
\midrule
DSSL & 57.43$\pm$3.51 & 18.60$\pm$2.25&25.68$\pm$3.73 & 26.15$\pm$0.46 & 0.76$\pm$0.10 & 1.27$\pm$0.17&44.70$\pm$2.44& 7.14$\pm$1.81 & 7.11$\pm$2.86 
 & 54.32$\pm$3.69 & 28.67$\pm$2.73 & 26.59$\pm$3.54, \\
HGRL &  61.97$\pm$3.10&  \underline{44.58$\pm$2.14} & \underline{37.05$\pm$4.78}&  \underline{29.79$\pm$1.11} & \underline{3.80 $\pm$0.83}&  \underline{4.09$\pm$1.16}&  \underline{60.56$\pm$3.72}&  \underline{44.61$\pm$3.32}&  \underline{35.65$\pm$5.24}&  61.14$\pm$1.49 & 34.06$\pm$1.71&  33.65$\pm$2.10\\
\name & \textbf{74.86$\pm$2.79} & \textbf{45.53$\pm$4.22} & \textbf{49.80$\pm$5.71} &\textbf{32.43$\pm$0.60} & \textbf{8.57$\pm$0.60} &\textbf{6.31$\pm$0.23} & \textbf{71.59$\pm$1.82} & \textbf{46.49$\pm$2.36} & \textbf{49.53$\pm$2.43} & \textbf{69.11$\pm$1.19} & \textbf{43.50$\pm$0.84} & \textbf{44.05$\pm$1.17}\\
\bottomrule
\bottomrule
\end{tabular}}
\end{table*}
(1) MLP achieves good results on four of six heterophilic graphs (Cornell, Texas, Wisconsin, Actor), indicating that node features only can play a vital role in node representation learning. On the other hand, the basic GCN model performs relatively well on the other two heterophilic graphs (Chameleon and Squirrel), suggesting that nodes in these two graphs require more contextual information from neighboring nodes. This further shows the presence of global diversity among different graphs.

(2) \name~ outperforms traditional supervised models (like GCN, GAT and MLP) and traditional self-supervised GNN models (like DGI, GMI, MVGRL, BGRL and GRACE). This is because these methods are designed for homophilic graphs without considering the heterophily in the graphs.
\begin{figure*}[t]
	\centering
	\subfloat[Texas]{\includegraphics[width=.4\columnwidth]{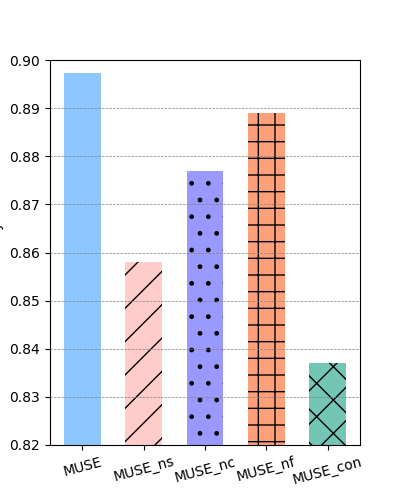}}
	\subfloat[Cornell]{\includegraphics[width=.4\columnwidth]{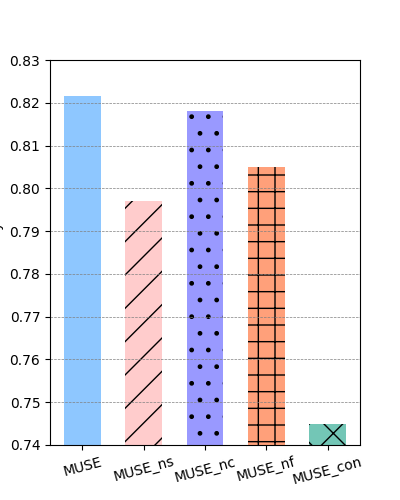}}
    \subfloat[Squirrel]{\includegraphics[width=.4\columnwidth]{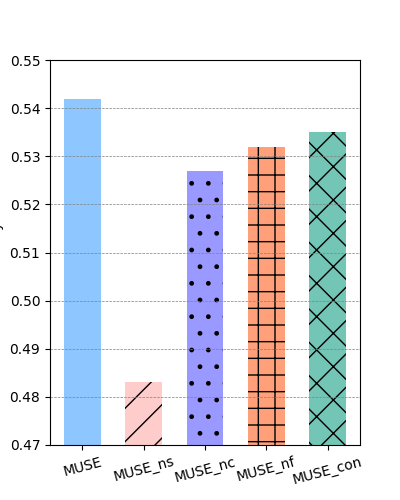}}
    \subfloat[Cora]{\includegraphics[width=.4\columnwidth]{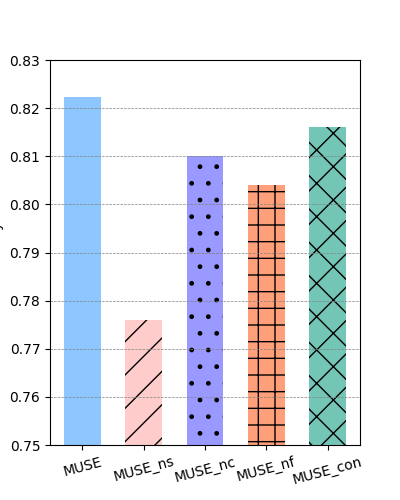}}\\
	\caption{Ablation study}
    \label{ablation}
\end{figure*}

(3) Existing unsupervised methods designed for heterophilic graphs exhibit a significant difference in performance across different datasets. DSSL and NWR-GAE perform even worse than MLP on four of the six heterophilic datasets (Cornell, Texas, Wisconsin and Actor), although performing well on the other two heterophilic datasets.
HGRL is the runner-up on Actor, but the accuracy on Chameleon is only 0.4829 ,while the best result is 0.7237 (\name). GREET ranks first on Cornell, but the accuracy score on Chameleon is 0.6364. We ascribe this instability in performance to the neglect of node diversity in these methods, which adopt a uniform processing strategy for all nodes and thus compromise the expressive power of the learned node representation. \name~ determines the amount of contextual information that each ego node needs in a node-specific way and shows superiority consistently on all six heterophilic datasets, ranking first on four of the datasets, and second on the other two.


(4) \name~ outperforms all the baselines in terms of average performance over all the datasets. On heterophilic graphs, the average performance of \name~  was higher than the runner-up which is a supervised method designed for heterophilic graphs, while there is a slight tradeoff in performance on homophilic graphs.

\subsubsection{Node Clustering Performance.}
The node clustering results on three heterophilic datasets and one homophilic dataset are reported in Table \ref{cluster}. From Table \ref{cluster}, we have the following observation:

It is notable that our model outperforms all the competitors on both heterophilic and homophilic datasets. We achieve relative improvement up to 19.22\% (ACC) on Texas than MVGRL. The improvement of ACC indicates that the node representation learned by our model can be correctly assigned to their respective clusters in the K-means clustering model. Relative improvement up to 125.53\% (NMI) on Actor than HGRL reflects a high degree of similarity between the clustering results of the node representation and true labels. Moreover, the 54.28\% (ARI) relative improvement on Actor than HGRL demonstrates excellent performance of our model in the clustering task even when considering the penalty for random clustering.

\subsection{Ablation Study}
In order to examine the effectiveness of each component of \name, we conduct ablation experiments on various variants of the model. We primarily validated the effectiveness of our model from two perspectives: (1) Effectiveness of the components in the node representation learning module, and (2) Effectiveness of the information fusion controller. 

\subsubsection{Effectiveness of the Components in the Node Representation Learning Module}
The major components in the node representation learning module include: semantic contrast, contextual contrast, and fusion contrast. 
To show the importance of each component,
we design model variants by removing different contrasts from our model.
Specifically,
we remove $\mathcal{L}_s$ from Eq.~\ref{contrast}
and
call the variant \textbf{MUSE\_ns}
(\textbf{n}o \textbf{s}emantic contrast);
we remove $\mathcal{L}_c$ and call 
the variant \textbf{MUSE\_nc} 
(\textbf{n}o \textbf{c}ontext contrast);
we call the variant removing $\mathcal{L}_r$ as \textbf{MUSE\_nf}(\textbf{n}o \textbf{f}usion contrast). We conduct experiments on three heterophilic graphs (e.g., Texas, Cornell and Squirrel) and one homophilic graph (e.g., Cora), the results of node classification task are shown in Figure \ref{ablation}. From the figure, 
we can see that:

(1) \name~ outperforms \name\_ns and \name\_nc on these four datasets. 
This shows that both semantic contrast and contextual contrast in the node representation learning module play a vital role.
Further,
the advantage of \name\ over \name\_nf shows that cross-view fusion contrast is also a necessity in enhancing the effectiveness of node representation learning.

(2) Compared with \name\_nc and \name\_nf, 
\name~ leads to a much larger performance gap than \name\_ns, 
especially on Squirrel and Cora. 
This 
highlights the importance of ego node's features for node representations in both homophilic and heterophilic graphs.

\subsubsection{Effectiveness of the Information Fusion Controller}

The information fusion controller takes into account the difference in the local diversity of node-neighborhood similarity across different nodes and the global distribution of similarity in the whole graph. With the information fusion controller, 
we can combine the semantic and contextual information at the node level to leverage the diversity of nodes. 
To validate its effectiveness, we set the value of $\lambda$ to 1 for each node and we call this variant as \textbf{MUSE\_con}(\textbf{con}troller). This variant takes the same amount of contextual information from the neighborhood for all nodes without considering the diversity among nodes.
Figure \ref{ablation} clearly indicates a decrease in performance of MUSE\_con compared with MUSE, especially on Texas and Cornell, suggesting the existence of diversity among nodes and the necessity of effectively combining semantic information and contextual information at the node level. Therefore, the information fusion controller has demonstrated a highly significant impact.

\subsection{Parameter Analysis}
In this section, we conduct experiments to investigate the impact of parameters in our model, including the pre-set hyper-parameter $\epsilon$ and the weight factors of contrastive loss $\beta_1$, $\beta_2$.

\subsubsection{Analysis of Weight Factor $\beta_1$ and $\beta_2$}
\begin{figure}[htbp]
    \centering
    \subfloat[Sensitivity of $\beta_1$]{\includegraphics[width=.5\columnwidth]{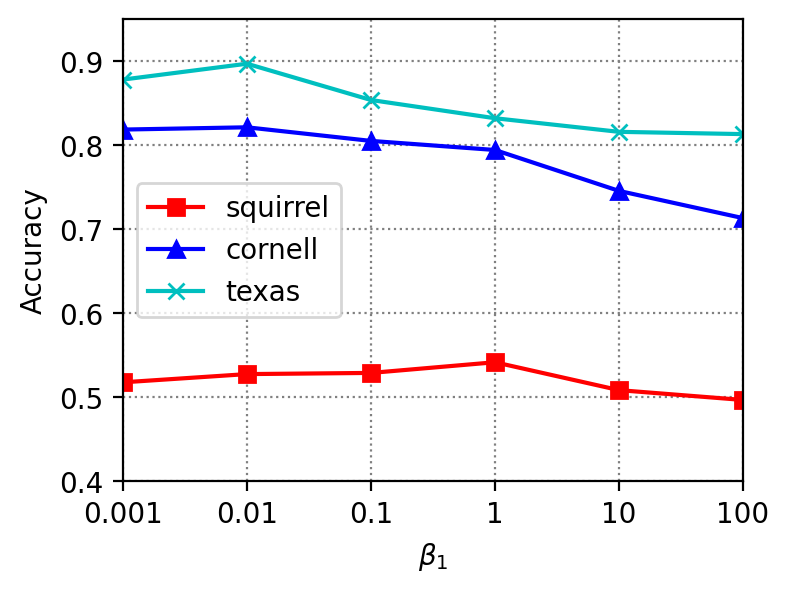}
    \label{b1}}
    \subfloat[Sensitivity of $\beta_2$]{\includegraphics[width=.5\columnwidth]{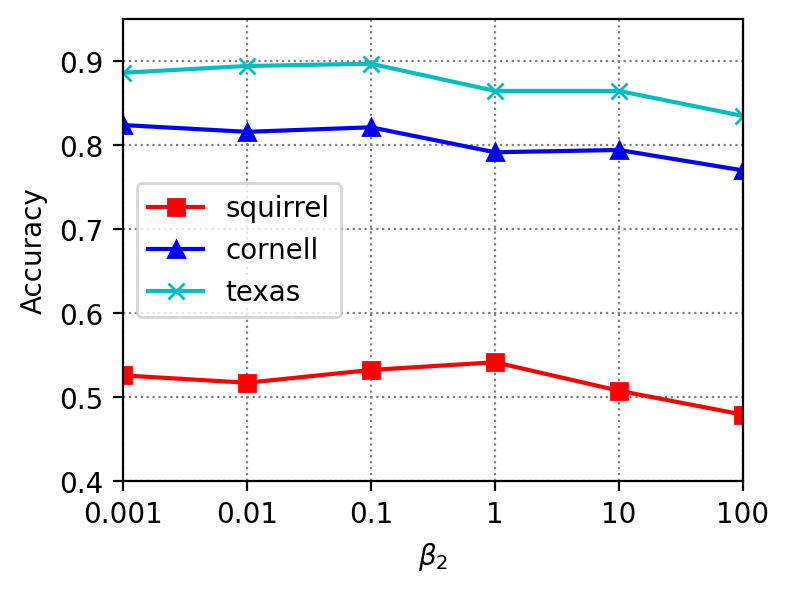}
    \label{b2}}\\
    \caption{Sensitivity analysis on weight factor $\beta_1$, $\beta_2$}
    \label{sensitivity}
\end{figure}

We study the sensitivity of our model with respect to the weight factors $\beta_1$ and $\beta_2$ in Equation \ref{contrast} by varying the values from $10^{-3}$ to $10^2$ (in Figure \ref{b1} and \ref{b2}). A common phenomenon is that a too large $\beta_1$ or $\beta_2$ leads to an obvious performance degradation, which indicates that excessive information from neighbors may have an opposite effect. Specifically, the best choice of $\beta_1$ and $\beta_2$ for Texas and Cornell is 0.01 and 0.1, as the features of ego node plays a more important role in this two datasets. The best choice of $\beta_1$ and $\beta_2$ for Squirrel is 1, indicating the importance of contextual information and fused information.
\subsubsection{Analysis of hyper-parameter $\epsilon$}
\begin{figure}[htbp]
  \centering
  \includegraphics[width=0.65\linewidth]{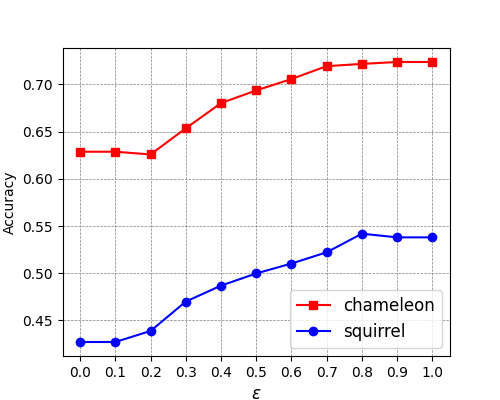}
  \caption{Sensitivity analysis on hyper-parameter $\epsilon$}
  \label{e}
\end{figure}

To study the impact of hyper-parameter $\epsilon$ on our model, we set $\epsilon$ from 0 to 1 and show the node classification accuracy in Figure \ref{e}.
We observe that the node classification accuracy is highly sensitive to the pre-set hyper-parameter on Squirrel and Chameleon. This suggests that our restriction on the global diversity of nodes in the graph is necessary. Specifically, performance on this two datasets is better with larger $\epsilon$, indicating that nodes in the graphs require more contextual information provided by neighboring nodes.

\subsection{Generalizability}
As previously discussed, labeled data is usually scarce in real-life scenarios. Therefore, we propose a self-supervised model on heterophilic graphs to alleviate the dependency on labels. 
By pre-training, we can obtain task-agnostic node representations and evaluate them on downstream tasks with a small quantity of labels, aiming to achieve superior performance.
In this section, we further investigate the effectiveness of the frozen representations obtained by pre-training with fewer downstream labels for fine-tuning. 

Except the commonly used training/validation/test partition of 48/32/20 adopted by most 
baselines on heterophilic graphs as in Section \ref{result}, we further follow the experimental settings in HGRL \cite{chen2022towards} and alter the data partition to 10/10/80. The performance on node classification task is given in Table \ref{Generalization ability}, and the results of baselines are derived from~\cite{chen2022towards}.
From the table, 
we can observe that, \name\ outperforms other competitors on Squirrel and Chameleon, 
and is the runner-up on Cora. 
This shows that \name\ 
still achieves impressive results on the downstream task despite the decrease in labeled data, particularly on heterophilic graphs, outperforming other unsupervised learning methods significantly. This further verifies that \name\ has strong generalization ability.

\begin{table}[htbp]
  \caption{Results of node classification task with few labels}
  \label{Generalization ability}
  \begin{tabular}{lcccc}
    \toprule
    \toprule
    Data & Methods & Squirrel & Chameleon & Cora \\
    \midrule
    X,A,Y & GCN & 39.50$\pm$1.54 & \underline{54.65$\pm$2.17} & 82.26$\pm$1.20\\
    X,A,Y & H2GCN & \underline{41.18$\pm$0.81} & 54.02$\pm$1.56 & 81.38$\pm$1.16\\
    \midrule
    X, A & MVGRL & 33.49$\pm$0.84 & 42.34$\pm$2.11 & \textbf{84.53$\pm$1.05}\\
    X, A & GRACE & 34.47$\pm$1.11 & 45.89$\pm$3.10 & 83.69$\pm$0.73\\
    X, A & BGRL & 31.50$\pm$0.57 & 45.54$\pm$1.94 & 83.01$\pm$0.71\\
    \midrule
    X, A & HGRL & 35.42$\pm$0.91 & 45.04$\pm$1.91 & 82.08$\pm$0.84\\
    X, A & \name &\textbf{41.67$\pm$0.90} & \textbf{57.89$\pm$1.27} & \underline{84.11$\pm$0.73}\\
  \bottomrule
  \bottomrule
\end{tabular}
\end{table}

\subsection{Efficiency}
In terms of efficiency, we evaluate the performance of our model against four state-of-the-art self-supervised learning baselines. Specifically, we focus on the training time during the representation pre-training stage and conduct experiment on Squirrel dataset to report the time cost for each epoch. The experimental results are shown in Table \ref{time}. 
It can be observed that \name\  requires significantly less training time per epoch compared to other three methods DSSL, NWR-GAE and GREET that also use GNN as the encoder.
While HGRL is more efficient,
it uses MLP as the encoder
and performs worse than \name\ in all the node classification/clustering comparisons as shown in Tables~\ref{classification} and~\ref{cluster}.
All these results show that our model is both effective and efficient.
\begin{table}[htbp]
  \caption{Comparison of efficiency on Squirrel}
  \label{time}
  \begin{tabular}{cccccc}
    \toprule
    \toprule
    Methods & DSSL & NWR-GAE & HGRL & GREET & \name\\
    \midrule
    Time & 1.848s & 1.104s & \textbf{0.11s} &1.136s & \underline{0.255s}\\
  \bottomrule
  \bottomrule
\end{tabular}
\end{table}

\section{Conclusion}
In this paper, we propose a novel model named \name~ for unsupervised node representation learning on heterophilic graphs. We employ GNNs enhanced with contrastive learning to capture the information from semantic and contextual views and fuse the node representations with an information fusion controller. Fusion contrast is utilized to enhance the effectiveness of the fused node representations. 
The information fusion controller considers both the local diversity of similarity across different ego nodes and the global distribution of similarity in the whole graph. 
We train the two components with an alternating strategy to boost each other. 
Extensive experiments reveal the effectiveness and generalizability of our method.


\bibliographystyle{ACM-Reference-Format}
\bibliography{sample-base}

\appendix

\end{document}